\def\paperlanguage{}
\pdfoutput=1 

\documentclass[letterpaper, 10 pt, conference]{ieeeconf}  

\usepackage{bm}
\usepackage{cite}
\usepackage{flushend}
\include{preamble}

\IEEEoverridecommandlockouts                              

\title{\LARGE \textbf
  {
    Modification of muscle antagonistic relations and hand trajectory on the dynamic motion of Musculoskeletal Humanoid
  }
}

\author{Yuya Koga, Kento Kawaharazuka, Moritaka Onitsuka, Tasuku Makabe, Kei Tsuzuki,\\
Yusuke Omura, Yuki Asano, Kei Okada and Masayuki Inaba
  \thanks{The authors are with the Department of Mechano-Informatics, Graduate School of Information Science and Technology, The University of Tokyo, 7-3-1 Hongo, Bunkyo-ku, Tokyo, 113-8656, Japan.
    {\texttt\small [koga, kawaharazuka, onitsuka, makabe, tsuzuki, oomura, asano, k-okada, inaba]@jsk.t.u-tokyo.ac.jp}
  }
}
\begin{document}

\maketitle
\thispagestyle{empty}
\pagestyle{empty}

\begin{abstract}
In recent years, some research on musculoskeletal humanoids is in progress.
However, there are some challenges such as unmeasurable transformation of body structure and muscle path, and difficulty in measuring own motion because of lack of joint angle sensor.
In this study, we suggest two motion acquisition methods.
One is a method to acquire antagonistic relations of muscles by tension sensing, and the other is a method to acquire correct hand trajectory by vision sensing.
Finally, we realize badminton shuttlecock-hitting motion of ``Kengoro'' with these two acquisition methods.
\end{abstract}

\section{INTRODUCTION} \label{sec:1}
These days, some research on musculoskeletal humanoids like \cite{ijars2013:nakanishi:approach}, \cite{2013:wittmeier:toward}, \cite{2013:jantsch:anthrob}, \cite{humanoids2016:asano:kengoro}, and \cite{iros2019:kawaharazuka:musashi} is in progress.
Musculoskeletal humanoids are created to imitate human body structures and their joints are driven by tendons corresponding to muscles.
Especially dynamic motion, like \cite{2010:niiyama:athlete} and \cite{2018:bledt:cheetah}, can lead to promotions of humanoid motion ability.

There are difficulties in dynamic motions of musculoskeletal humanoids.
They are caused by unmeasurable body deformation and tendon path change.
These difficulties result in differences between a kinematic model and a real robot and make it difficult to control musculoskeletal humanoids accurately.
At the same time, error in relations of muscles results in high internal tension and overheat.

Although there are several kinds of research on these difficulties, there are still problems.
In a previous study \cite{humanoids2013:asano:motion}, load sharing of muscle were attempted and the method enabled low tension movement of Kenshiro \cite{humanoids2012:nakanishi:kenshiro}.
But in this research, muscles which are applied load sharing were determined by a human in advance so it cannot apply to another motion or more complex motion.
A tension-based joint-space controller for musculoskeletal humanoids was studied in \cite{humanoids2016:kawamura:joint} \cite{jantsch2011scalable} \cite{jantsch2012computed}.
The tension-based controller enabled complex multiple degrees of freedom (DOF) joint, head and neck joint of Kengoro \cite{humanoids2016:asano:kengoro}, but the difference between a kinematic model and a real robot is still a challenge.

Antagonist inhibition control (AIC), which loosen antagonist muscles completely, was developed in \cite{iros2017:kawaharazuka:antagonist}.
AIC realized wide range motions in scapula and arm of Kengoro.
In addition, online joint muscle mapping acquisition using vision was developed in \cite{icra2018:kawaharazuka:online} \cite{kawaharazuka2018bodyimage} \cite{kawaharazuka2019longtime} and enabled accurate hand position control of Kengoro in a real environment.
However, dynamic motion with a target object is still difficult because of large hysteresis and control delay.

\begin{figure}[t]
 \begin{center}
  \includegraphics[width=1.0\columnwidth]{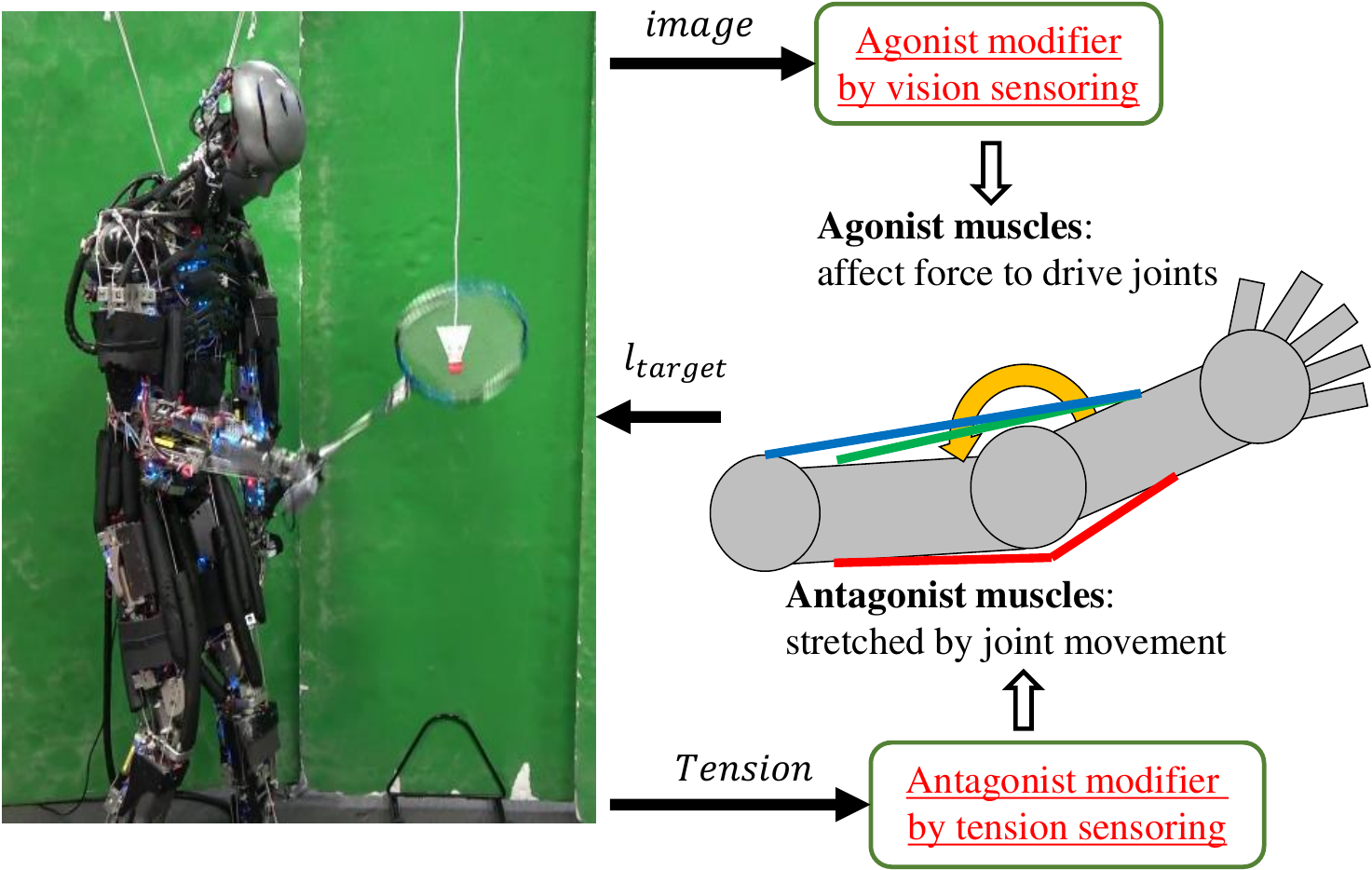}
  \caption{System overview of this research.}
  \label{fig:overview}
 \end{center}
\end{figure}

\begin{figure}[t]
  \centering
  \includegraphics[width=0.75\columnwidth]{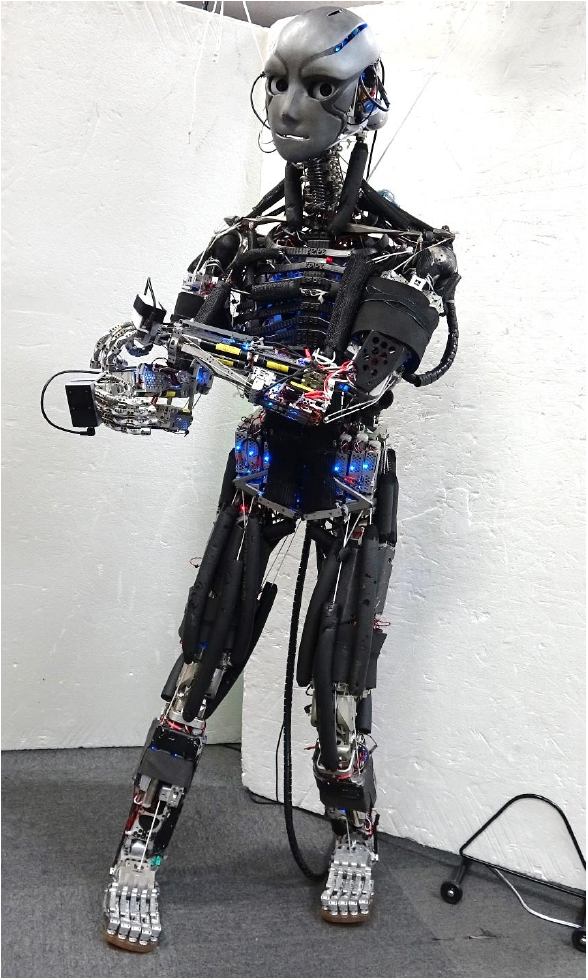}
  \caption{Musculoskeletal humanoids Kengoro\cite{humanoids2016:asano:kengoro}.}
  \label{fig:kengoro_introduction}
\end{figure}

The goal of this research is to tackle these problems with two types of control methods shown in \figref{fig:overview}.
One is agonist modifier applied to agonist muscles which give force to drive robot joints.
Agonist modifier enables a robot to realize objective motion.
The other is antagonist modifier applied to antagonist muscles which are stretched by joint movement.
Antagonist modifier decreases high internal forces and enables smooth dynamic motion of musculoskeletal humanoid.
Finally, by adding two control methods we realized badminton hitting motion of Kengoro \figref{fig:kengoro_introduction} which is developed in our group.


This paper is composed as follows.
In \secref{sec:1}, we stated background and goal of this research.
In \secref{sec:2}, we will propose two control methods, antagonist modifier and agonist modifier.
In \secref{sec:3}, we will show two experiments which evaluate these proposing methods.
Finally, in \secref{sec:4}, we will give a conclusion and future works of this research.

\section{ANTAGONIST MODIFIER AND AGONIST MODIFIER} \label{sec:2}
\subsection{Antagonist modifier by tension sensing}
\begin{figure}[t]
 \begin{center}
  \includegraphics[width=1.0\columnwidth]{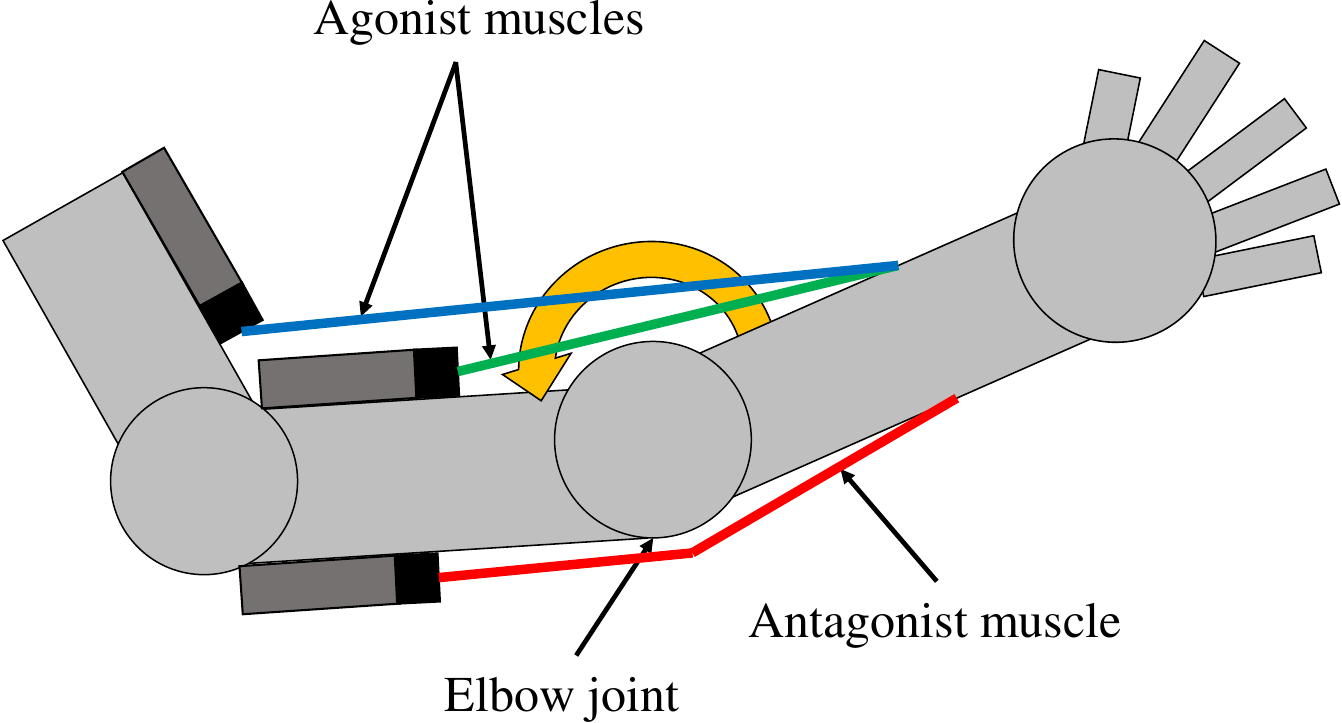}
  \caption{Figure of Agonist muscle and Antagonist muscle.}
  \label{fig:agonist_antagonist}
 \end{center}
\end{figure}

Muscles are classified into two types, agonist muscles and antagonist muscles.
In musculoskeletal humanoids, one joint is driven by multiple muscles like the human body.
Those types are defined in relation to joint move direction, that is, for example, yellow arrow in \figref{fig:agonist_antagonist}.
Agonist muscles give force to drive joints in the desired direction.
In \figref{fig:agonist_antagonist}, blue muscle and green muscle are agonist muscles.
It means that elbow joint is driven in the direction to the yellow arrow when these muscles pull wires.
Antagonist muscles are stretched by joint movements.
In \figref{fig:agonist_antagonist}, red muscle is an antagonist muscle.
It means that this muscle is stretched when elbow joint moves in the direction to yellow arrow.

Antagonist modifier can be paraphrased as suppression of antagonist muscles.
In an ideal robot model, antagonist muscles follow joint movement accurately.
It means that tension of antagonist muscles are always zero and they have no looseness.
However, in the real robot, differences with the ideal robot model cause too much tension of antagonist muscles.
This tension opposes muscle movement and higher tension of agonist muscles.
So, by measuring tensions of antagonist muscles, antagonist modifier modifies muscle length path to suppress antagonist muscles.

\begin{figure}[t]
 \begin{center}
  \includegraphics[width=1.0\columnwidth]{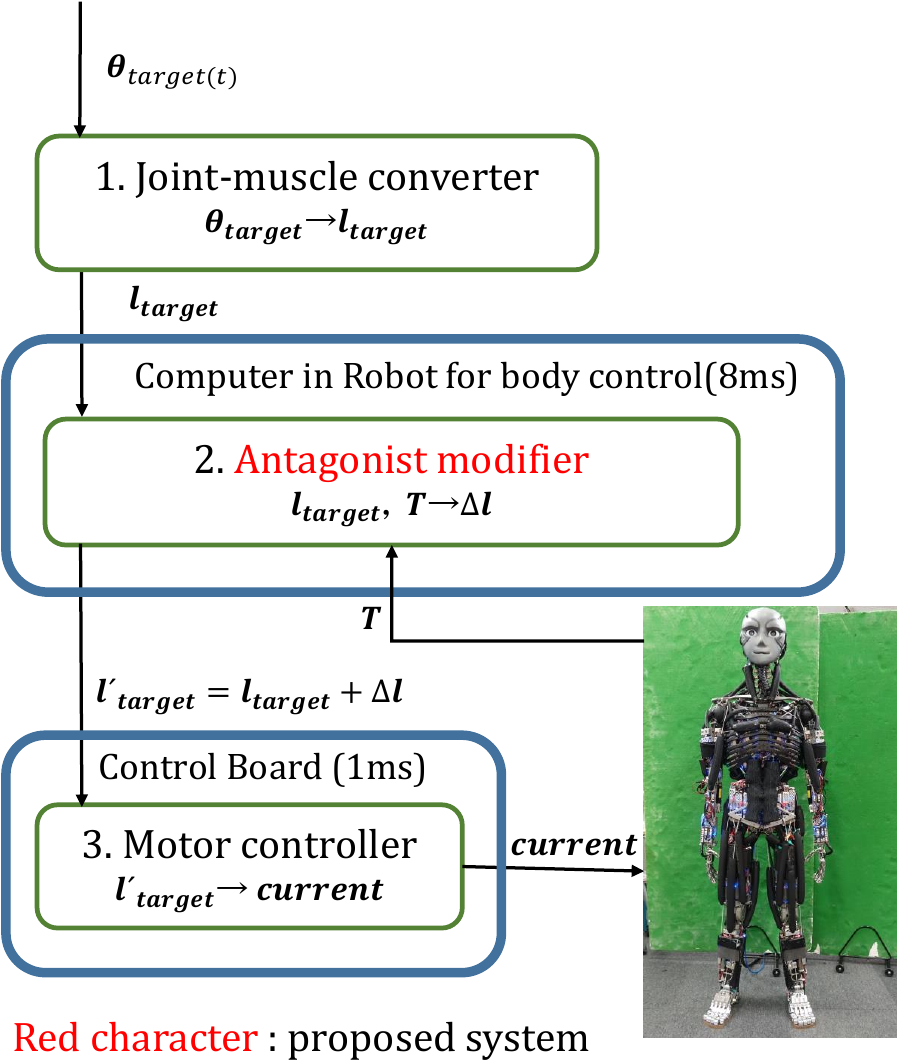}
  \caption{System of antagonist modifier.}
  \label{fig:system2}
 \end{center}
\end{figure}

The whole system of antagonist modifier is shown in \figref{fig:system2}.
We explain each element in the whole system one by one.

\begin{description}
\item [1. Joint-muscle convertor]\mbox{}

This element converts from reference joint angles to reference muscle lengths by using a kinematic robot model.
The kinematic model includes joint structures and tendon path information.
First reference joint angles are reflected in the kinematic model, and then reference muscle lengths are calculated kinematically.

\item [2. Antagonist modifier]\mbox{}

This element is the proposing method.
Calculation flow is as follows.
\begin{enumerate}
  \item Receive reference muscle lengths by Joint-muscle converter
  \item First Motion: send those muscle lengths to the real robot
  \item Measure muscle tensions while moving, and calculate correction factors of antagonist muscle lengths
  \item Second Motion: send reference muscle lengths with calculated correction factors
\end{enumerate}

Correction factors are calculated while sending reference muscle lengths.
Those factors are used in the next motion.

Next, we will state the calculation of correction factors.
Muscles are sorted into agonist muscles and antagonist muscles by referring to reference muscle lengths.
About antagonist muscles, correction factors are calculated as follows.
\begin{equation}
\Delta {l}_{antagonist}[i] = C * max(T[i] - {T}_{threshold}, 0)
\end{equation}
$\Delta {l}_{antagonist}[i]$ is correction factor of the $i$-th muscle, and $C$ is coefficient, and $T[i]$ is tension of the $i$-th muscle, and ${T}_{threshold}$ is threshold of muscle tension.
${T}_{threshold}$ is necessary because musculoskeletal robot cannnot keep its pose without minimum muscle tension.
In the experiment, we used $1kg$ as ${T}_{threshold}$.
These correction factors are calculated and recorded in each cycle.
At the end of one motion, those factors are smoothed with moving average filter.
In the next motion, reference muscle lengths added $\Delta {l}_{antagonist}$ are sent to the real robot.

\item[3. Motor controller]\mbox{}

This element controls motors by referential muscle length.
This control was developed in \cite{robio2011:shirai:whole}.
Calculation of target tension is as follows.
\begin{equation}
{T}_{target}[i] = max \{ 0, k[i] (L[i] - {L}_{ref}[i]) \} + {T}_{offset}[i]
\end{equation}
$k[i]$ is coefficient, and $L[i]$ is muscle length of $i$-th muscle, and ${L}_{ref}[i]$ is reference muscle length of $i$-th muscle.
The first member of the right side, $max \{ 0, k[i] (L[i] - {L}_{ref}[i]) \}$, is corresponds to one side spring and gives force in direction to only shrinking.
The second member of the right side, ${T}_{offset}[i]$, corresponds to tension offset and gives minimum force not to loosen too much.

Finally, FPGA motor controller calculates target current of each motor.
\end{description}

\subsection{Agonist modifier by vision sensing}
Musculoskeletal humanoids have difficulties in dynamic motion with target objects.
The error between the kinematic robot model and the real robot causes the error of hand trajectory.
So we have to modify hand trajectory by recognizing real hand trajectory and target position.

\begin{figure}[t]
  \begin{minipage}{0.80\hsize}
    \begin{center}
      \includegraphics[width=0.8\columnwidth]{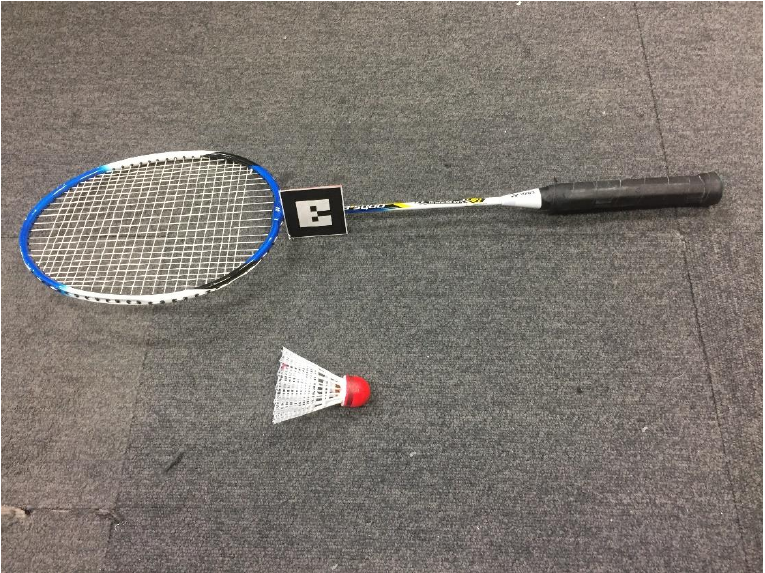}
      \caption{Shuttle and racket.}
      \label{fig:shuttle_racket}
    \end{center}
  \end{minipage}
  \begin{minipage}{0.80\hsize}
    \begin{center}
      \includegraphics[width=1.0\columnwidth]{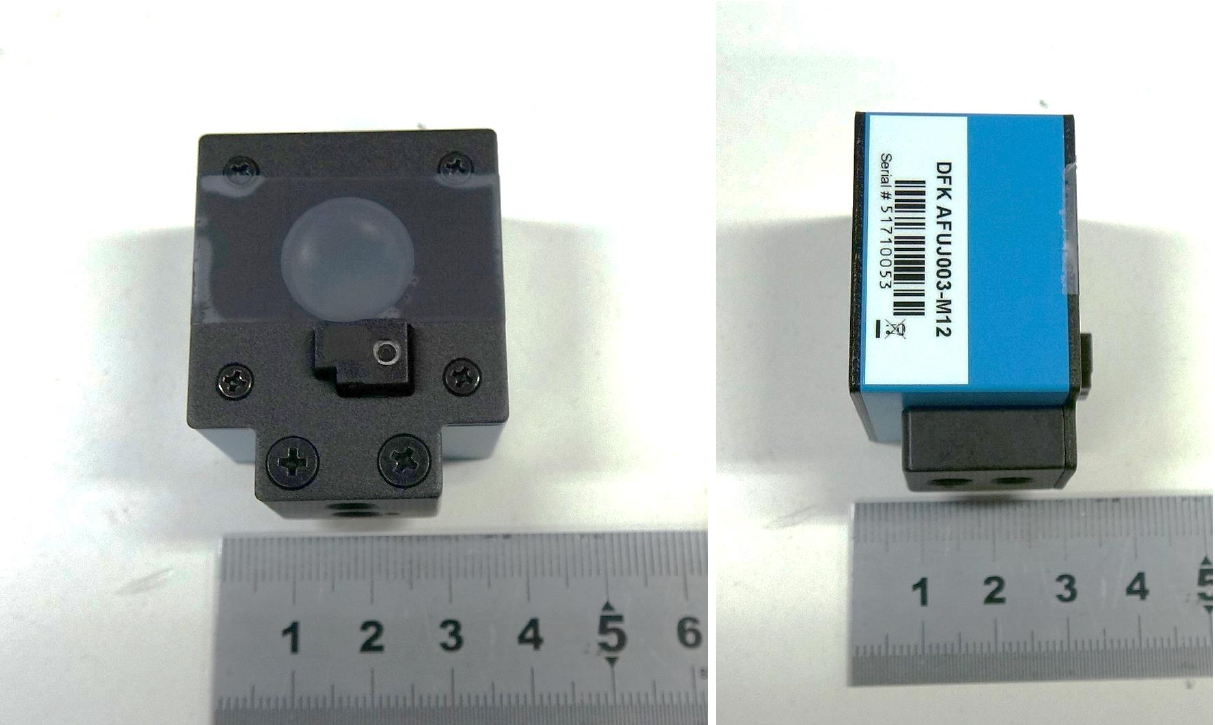}
      \caption{Camera with lens.}
      \label{fig:camera_picture}
    \end{center}
  \end{minipage}
\end{figure}

Through acquiring badminton hitting motion of Kengoro, we will show the effectiveness of agonist modifier.
Target shuttlecock and racket are shown in \figref{fig:shuttle_racket}.
Kengoro recognizes racket position, instead of hand trajectory, and target shuttlecock with two cameras shown in \figref{fig:camera_picture} mounted on his eye.

\begin{figure}[tp]
 \begin{center}
  \includegraphics[width=1.0\columnwidth]{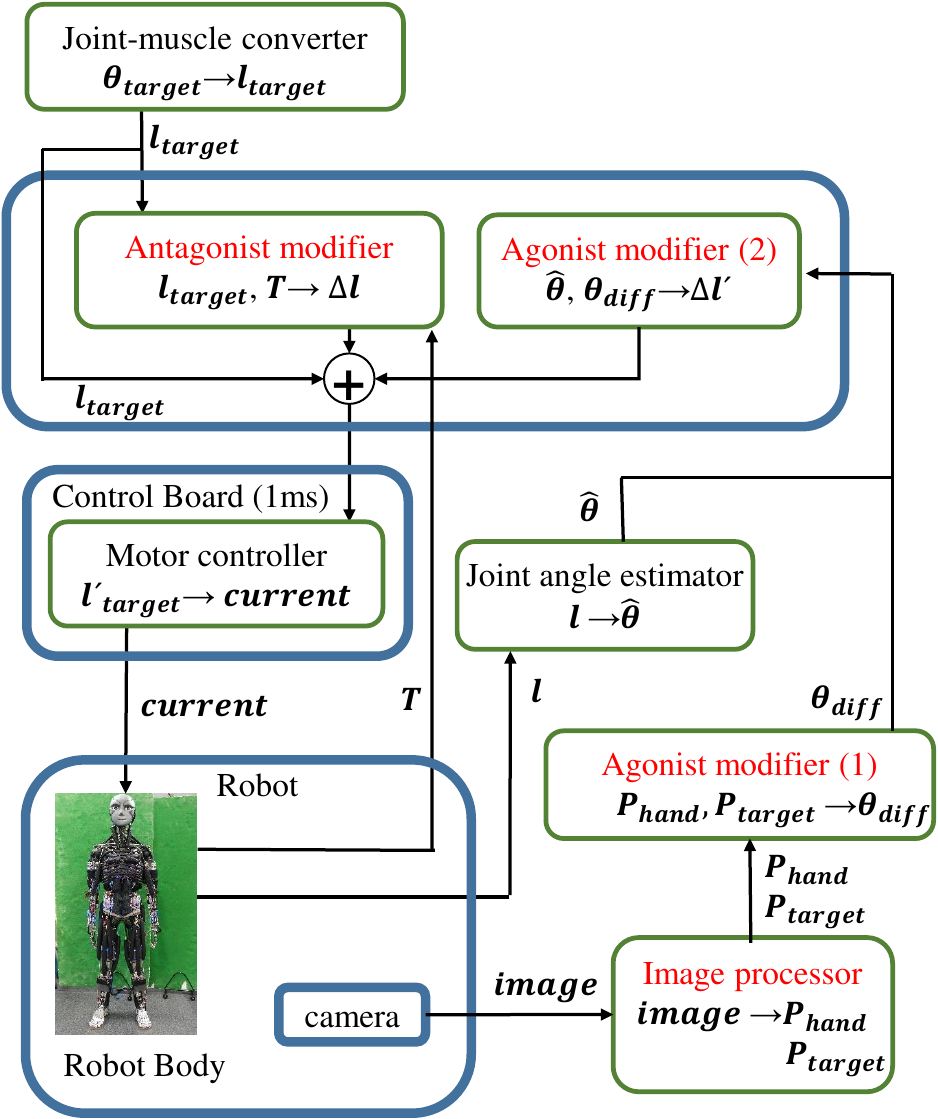}
  \caption{System of agonist modifier and antagonist modifier.}
  \label{fig:system3}
 \end{center}
\end{figure}

The whole system is shown in \figref{fig:system3}.
First, we will state image processing of this system and second we will state agonist modifier calculation.

\subsubsection{Image Processing}

Abstract of cameras used in this research is as follows.
Two cameras shown in \figref{fig:camera_picture} are mounted on Kengoro's eye.
Camera specifications are mentioned in \tabref{table:kengoro_camera_spec}.
In this research, resolution \& FPS mode is 1920$\times$1080 \& 30 FPS, and images were compressed to 960$\times$540 by thinning out pixels.

\begin{table}[htb]
    \begin{center}
        \caption{Basic specifications of camera on Kengoro \cite{humanoids2018:makabe:evelopment}.}
            \begin{tabular}{|c|c|} \hline
                camera size[mm] & 35$\times$35$\times$40 \\ \hline
                sensorsize & 1/2.3 \\ \hline
                focus & autofocus \& changable \\ \hline
                focus length[mm] & 5.4 \\ \hline
                lens & low distortion mini lens \\ \hline
                 & 3872$\times$2764, 7 \\ \cline{2-2}
                resolution \& FPS & 1920$\times$1080, 30 \\ \cline{2-2}
                 & 640$\times$480, 90 \\ \hline
            \end{tabular}
        \label{table:kengoro_camera_spec}
    \end{center}
\end{table}

\begin{figure}[t]
 \begin{center}
  \includegraphics[width=1.0\columnwidth]{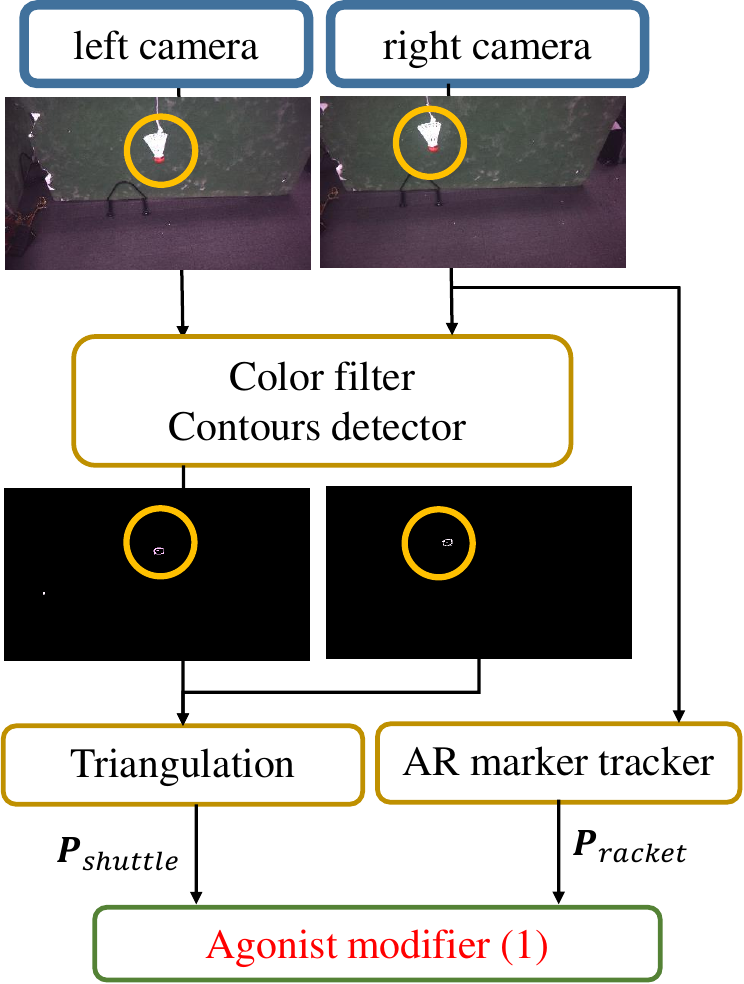}
  \caption{Flow of image processing.}
  \label{fig:image_recog_flow}
 \end{center}
\end{figure}

The flow of image processing is shown in \figref{fig:image_recog_flow}.
Racket position is recognized by position detection of AR marker mounted under the racket face.
This is because Kengoro have no angle sensor thus it is difficult to calculate accurate racket position without AR marker.
Target shuttlecock position is recognized by triangulation with two cameras.
In triangulation, we created pinhole camera models of each camera and calculated the 3-dimensional position based on the fixed positional relation between two cameras.

\subsubsection{Agonist modifier}

Based on the results of image processing, we calculated agonist modifier.
The Calculation method is divided into two parts.

In the first part, agonist modifier (1), correction factors of joint angles are calculated by \equref{eq:agonist_modifier_0} and \equref{eq:agonist_modifier_1}.
\begin{equation}
{\bm{P}}_{diff} = {\bm{P}}_{shuttle} - {\bm{P}}_{racket} \label{eq:agonist_modifier_0}
\end{equation}
\begin{equation}
{\bm{\theta}}_{diff} = {\bm{J}}^{\#} {\bm{P}}_{diff} \label{} \label{eq:agonist_modifier_1}
\end{equation}
${\bm{P}}_{shuttle}$ is shuttlecock position, and ${\bm{P}}_{racket}$ is racket position, and ${\bm{P}}_{diff}$ is the difference between these two positions, $\bm{J}$ is Jacobian between joint angle and hand position, ${\bm{\theta}}_{diff}$ is correction factors of joint angles.
${\bm{J}}^{\#}$ means pseudo-inverse of $\bm{J}$.
In this research, we considered correction factors of the elbow joint angle.

In the second part, agonist modifier (2), we convert correction factors of joint angles to correction factors of muscle lengths, $\Delta {\bm{l}}^{\prime}$, by using joint-muscle Jacobian $\bm{G}(\hat{\bm{\theta}})$ as shown in \equref{eq:agonist_modifier_2}.
\begin{equation}
\Delta {\bm{l}}^{\prime} = \bm{G}(\hat{\bm{\theta}}) {\bm{\theta}}_{diff} \label{eq:agonist_modifier_2}
\end{equation}
Joint-muscle Jacobian describes relations between joint angles and muscle lengths, and it is calculated by estimated joint angles $\hat{\theta}$ and kinematic robot model.
Joint angles are estimated by nonlinear muscle-joint state mapping developed in \cite{humanoids2015:okubo:muscle-learning} and \cite{kawaharazuka2018estimator}.
Finally, those correction factors are added to only agonist muscles in order not to compete with antagonist modifier.

\section{EXPERIMENTS} \label{sec:3}
In this section, we will show you two experiments to evaluate proposing methods.
In the first experiment, we evaluated antagonist modifier by executing dynamic hand movements and measuring tension transitions.
In the second experiment, we evaluated both agonist modifier and antagonist modifier by realizing badminton hitting motion of Kengoro.

\subsection{Antagonist modifier Experiment}
In this experiment, 13 degrees of freedom (DOF) were used.
6 DOFs were in right scapula and 7 DOFs were in right shoulder, elbow, and wrist.
Kengoro executed the same motion twice and we measured transitions of muscle lengths and muscle tensions.
After that, we compared these transition between the first motion and the second motion.

\begin{figure}[t]
 \begin{center}
  \includegraphics[width=1.0\columnwidth]{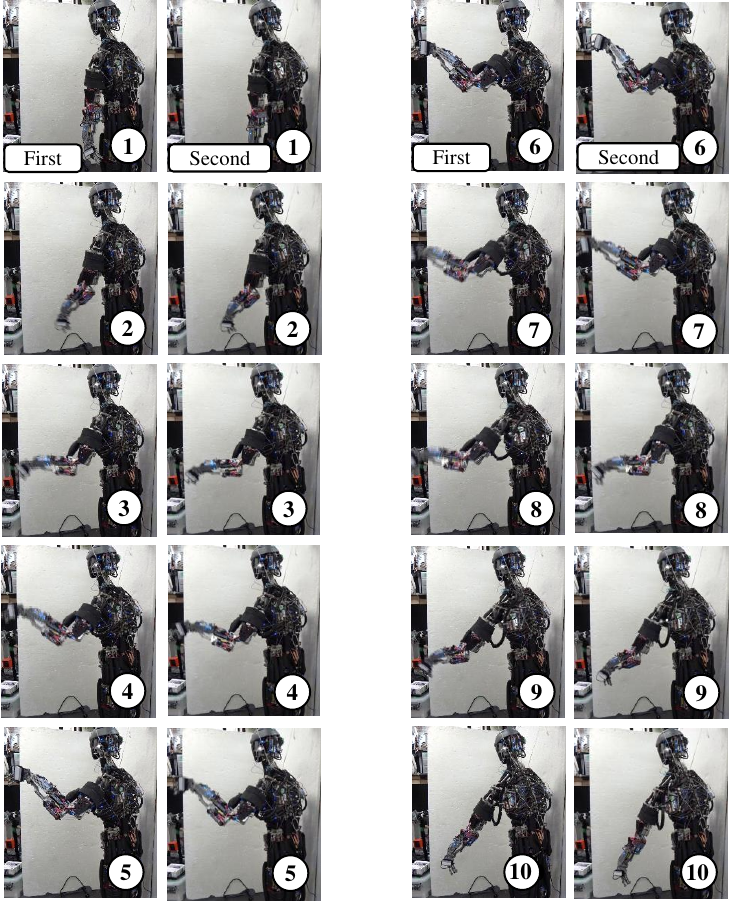}
  \caption{Experiment to assess antagonist modifier.}
  \label{fig:expt2}
 \end{center}
\end{figure}

The flow of the experiment is as follows.
\begin{enumerate}
\item Execute offline planned motion
\item Measure muscle tension transition and modify antagonist muscle length trajectory
\item Execute motion modified by antagonist modifier
\end{enumerate}
Actual robot motion is shown in \figref{fig:expt2}.

\begin{figure}[tp]
 \begin{center}
  \includegraphics[width=1\columnwidth]{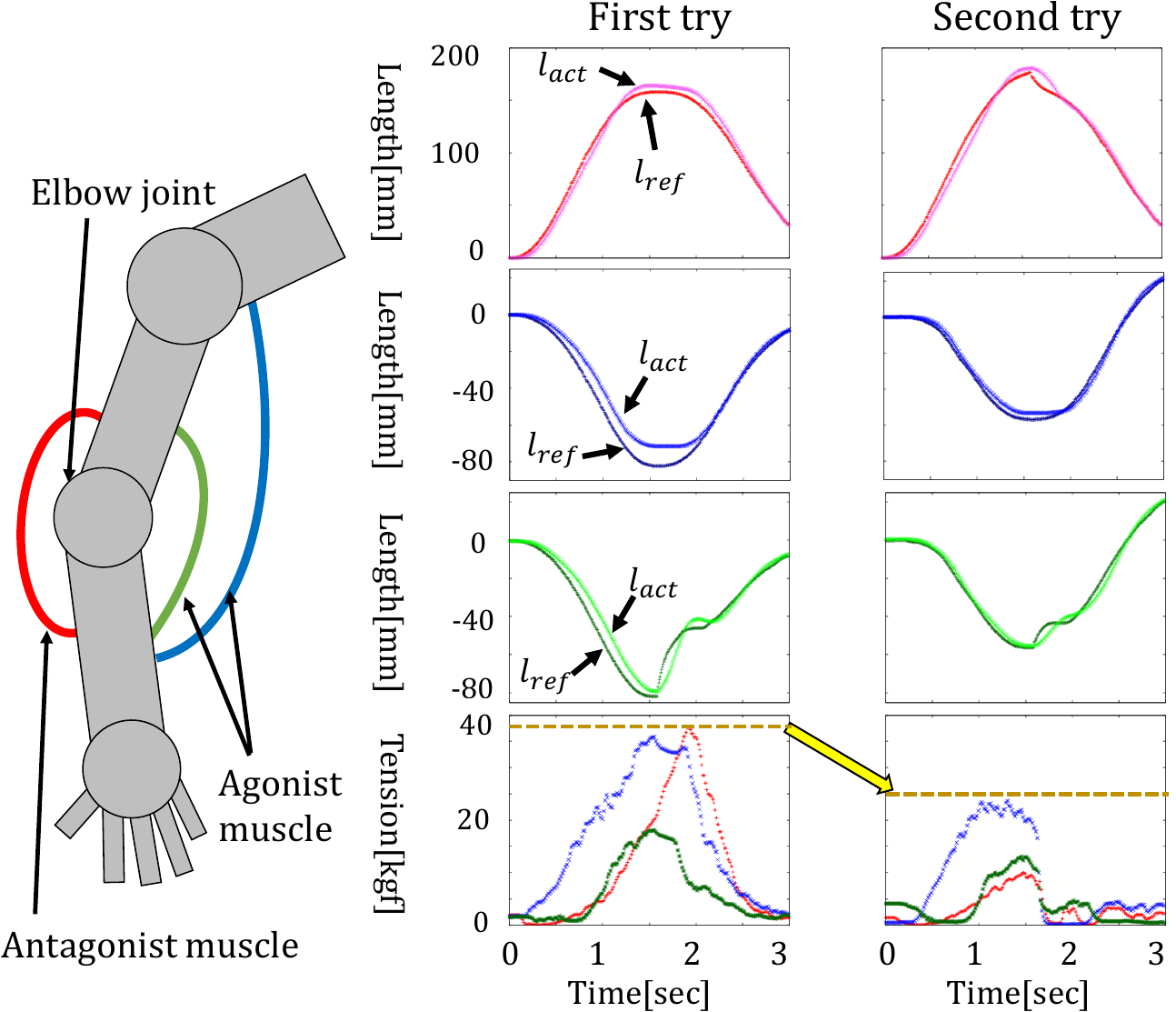}
  \caption{Transition of muscle length and tension.}
  \label{fig:expt2_graph}
 \end{center}
\end{figure}

\figref{fig:expt2_graph} shows comparisons between reference muscle lengths and real muscle lengths, and muscle tensions.
Each curved line in the left figure describes a muscle and its color corresponds to the colors in the right graphs.
In the first half of this motion, red muscle is an antagonist muscle, and agonist muscles are green and blue muscles.

As shown in tension transition graph, a tension of antagonist muscle, red muscle, greatly decreased in the second motion compared with the first motion.
As a result, tensions of agonist muscles, blue and green muscles, also decreased.
It means that internal force was decreased and load reduction was achieved.

\subsection{Agonist and Antagonist modifier Experiment}
The flow of the experiment is as follows.
\begin{enumerate}
\item Execute first swing motion which is planned offline previously \\
      We made this motion by only kinematic robot model without AIC.
\item Apply agonist modifier and modify hand trajectory\\
      In this part, the transitions of only agonist muscles are modified.
\item Execute second swing motion modified by agonist modifier
\item Apply antagonist modifier and modify antagonistic relations of muscles \\
      In this part, the transitions of only antagonist muscles are modified.
\item Execute third swing motion modified by both agonist modifier and antagonist modifier
\end{enumerate}

The motion by only kinematic robot model without AIC means that it does not consider errors between the kinematic model and the real robot.
Therefore, the real robot must have too much internal force when executing that initial motion.

\begin{figure}[t]
 \begin{center}
  \includegraphics[width=1.0\columnwidth]{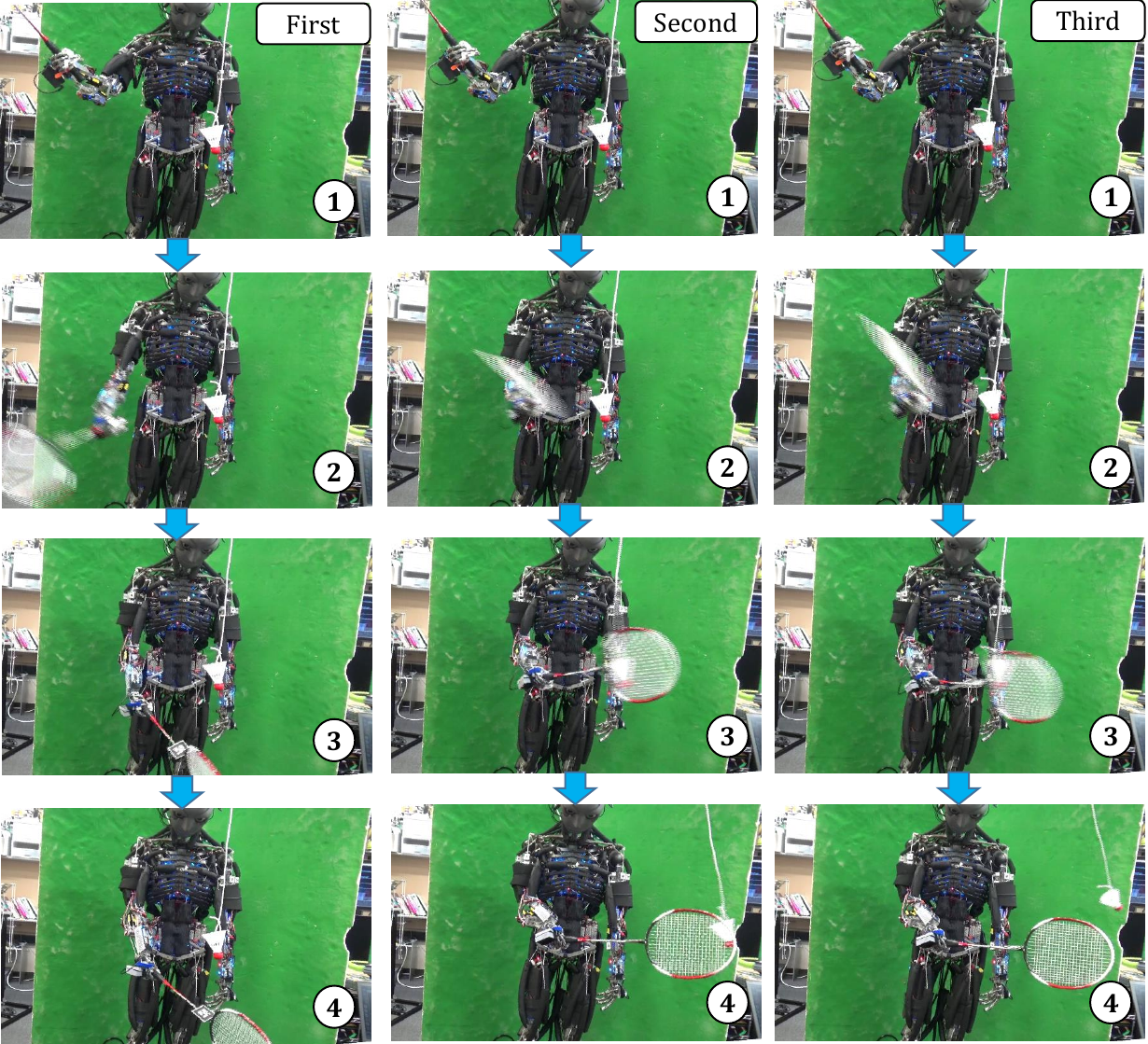}
  \caption{Experiment of both agonist modifier and antagonist modifier.}
  \label{fig:expt3}
 \end{center}
\end{figure}

Actual robot motion is shown in \figref{fig:expt3}.
The first line shows the previously planned motion, and the second line shows the agonist-modifier-applied motion, and the third line shows both agonist and antagonist-modifier-applied motion.

By applying agonist modifier, it is shown that Kengoro acquired badminton hitting motion.


\section{CONCLUSION} \label{sec:4}
In this research, we realized badminton hitting motion of musculoskeletal humanoid Kengoro by applying two control methods, antagonist modifier and agonist modifier.
The dynamic motion of musculoskeletal humanoids has several difficulties such as hand trajectory error and muscle antagonistic relation error.
To tackle hand trajectory error, we proposed agonist modifier in which the robot recognizes his hand trajectory and modifies his motion automatically.
To tackle muscle antagonistic relation error, we proposed antagonist modifier which suppresses tension of antagonist muscles and reduces internal force.
Next, we showed two experiments to evaluate these two methods.
In the first experiment, antagonist modifier enables Kengoro to move smoothly with low internal force.
In the second experiment, both agonist modifier and antagonist modifier enables Kengoro to acquire badminton hitting motion.

In the future, we would like to apply these methods to the whole body and improve the hardware of Kengoro to realize wider active degrees of freedom.
To apply to the whole body, we have to consider transitions of muscle type, agonist muscle to antagonist muscle and antagonist muscle to agonist muscle.
These transitions make it difficult to move a lot of muscles in more complicated movements.
In addition to software improvement, we would like to improve the hardware of the robot to realize wider active movable range.

Another future work is realizing the reusability of these two acquisition methods.
In this paper, correction terms of Agonist modifier and Antagonist modifier cannot apply to other motions.
If we can record these correction terms in a more abstract form, we can apply correction terms to other motions.

\section*{ACKNOWLEDGMENT}
I am deeply grateful to support from TOYOTA MOTOR CORPORATION.

{
  \bibliographystyle{IEEEtran}
  \bibliography{main}
}

\end{document}